\documentclass[10pt,twocolumn,letterpaper]{article}

\usepackage[pagenumbers]{cvpr}      %

\usepackage{graphicx}
\usepackage{amsmath}
\usepackage{amssymb}
\usepackage{booktabs}
\usepackage{multirow}
\usepackage{marvosym}

\usepackage[pagebackref,breaklinks,colorlinks]{hyperref}

\usepackage[capitalize]{cleveref}
\crefname{section}{Sec.}{Secs.}
\Crefname{section}{Section}{Sections}
\Crefname{table}{Table}{Tables}
\crefname{table}{Tab.}{Tabs.}

\def\cvprPaperID{3029} %
\def\confName{CVPR}
\def\confYear{2023}

\begin{document}

\title{ClassPruning: Speed Up Image Restoration Networks by Dynamic \emph{N:M} Pruning}

\renewcommand{\thefootnote}{\fnsymbol{footnote}}
\author {
	Yang Zhou{\footnotemark[2]}
	\quad Yuda Song{\footnotemark[2]}
	\quad Hui Qian
	\quad Xin Du{\textsuperscript \Letter}\\
	Zhejiang University, Hangzhou, China \\
	{\tt\small \{yang\_zhou,syd,qianhui,duxin\}@zju.edu.cn}
}

\maketitle

\footnotetext[2]{Equal contribution.}
\renewcommand{\thefootnote}{\arabic{footnote}}

\begin{abstract}
Image restoration tasks have achieved tremendous performance improvements with the rapid advancement of deep neural networks.
However, most prevalent deep learning models perform inference statically, ignoring that different images have varying restoration difficulties and lightly degraded images can be well restored by slimmer subnetworks.
To this end, we propose a new solution pipeline dubbed ClassPruning that utilizes networks with different capabilities to process images with varying restoration difficulties.
In particular, we use a lightweight classifier to identify the image restoration difficulty, and then the sparse subnetworks with different capabilities can be sampled based on predicted difficulty by performing dynamic $N$:$M$ fine-grained structured pruning on base restoration networks.
We further propose a novel training strategy along with two additional loss terms to stabilize training and improve performance. 
Experiments demonstrate that ClassPruning can help existing methods save approximately 40\% FLOPs while maintaining performance.
\end{abstract}
\vspace{-0.2cm}
\section{Introduction}

\begin{figure}[t]
    \centering
    \includegraphics[width = 0.47\textwidth]{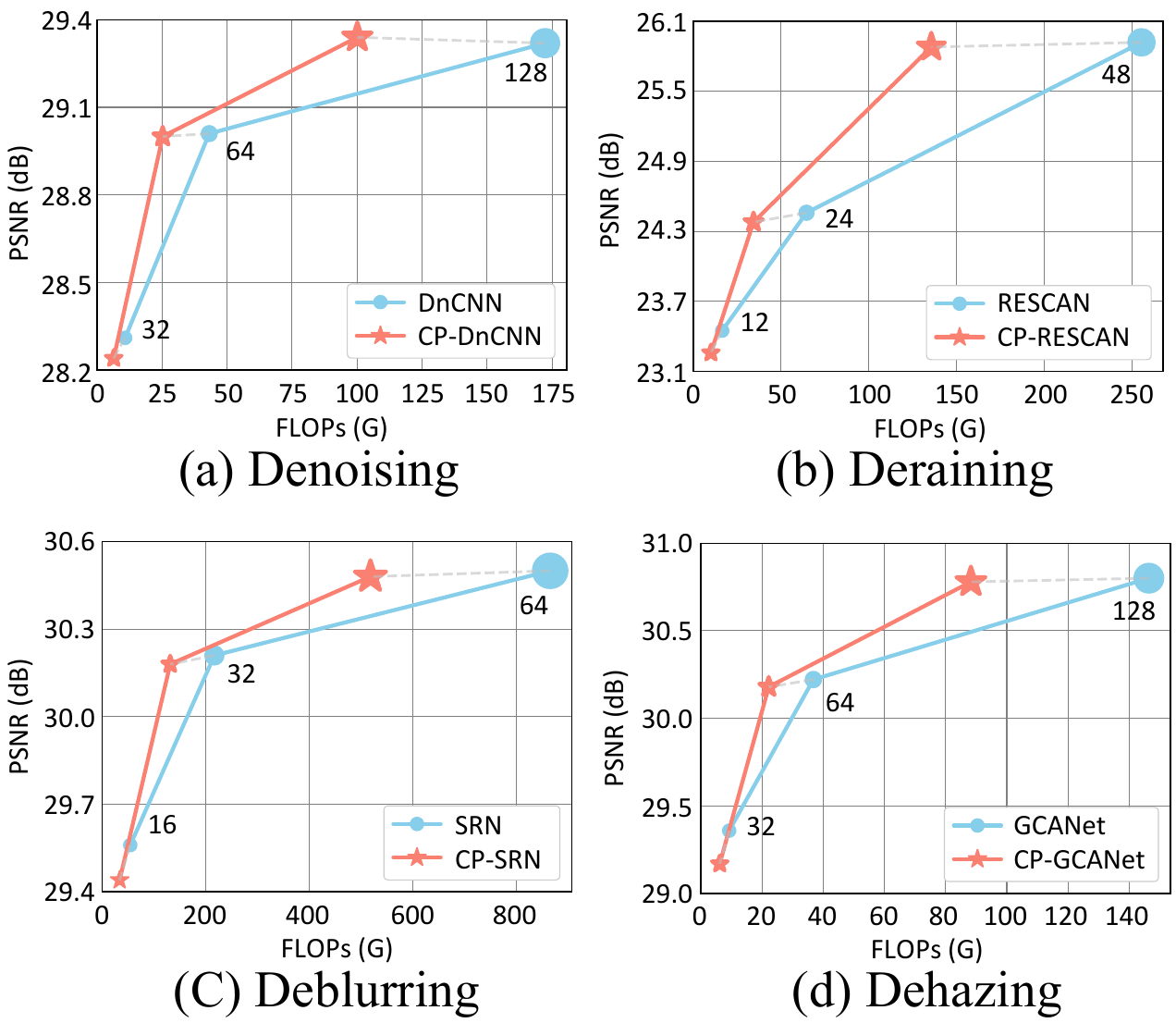}
    \centering
    \caption{Our ClassPruning can achieve comparable performance and reduce approximately 40\% computational costs under different network widths for various image restoration tasks.}
    \label{Fig.Front}
    \vspace{-0.22cm}
\end{figure}

Image restoration aims to reconstruct high-quality clean images from low-quality inputs corrupted by various degradations. 
Due to the ill-posed nature, obtaining the optimal result in the infinite solution space is highly challenging.
Recently, deep neural networks (DNNs) have developed rapidly and gradually become a preferred alternative to conventional restoration methods.
Its strong capability to restore degraded images has been witnessed in various image restoration tasks, including image denoising~\cite{wang2022uformer,zamir2022restormer}, deraining~\cite{purohit2021spatially,zamir2021multi}, deblurring~\cite{zhang2019deep,cho2021rethinking}, dehazing~\cite{qin2020ffa,ye2021perceiving}, etc.

Despite the remarkable success, high-capacity image restoration networks tend to be quite complex, resulting in significant inference costs, especially when used on mobile devices with limited computational resources.
Therefore, in addition to performance, the computational cost is also a crucial enabling factor~\cite{szegedy2016rethinking}.
And researchers have investigate various methods such as knowledge distillation~\cite{hinton2015distilling}, quantization~\cite{jacob2018quantization,zhou2017incremental}, designing efficient model architecture~\cite{howard2017mobilenets} and pruning~\cite{wen2016learning,li2016pruning} to speed up DNNs.
However, these methods usually obtain a static network applied to all input samples, which limits the model representation capability, as the diverse demands for network parameters and capacity from different instances are neglected.
Dynamic networks, as opposed to static ones, are proposed to adapt their structures or parameters to the input during inference to enjoy favorable properties absent in static models.
It uses lightweight networks to deal with easy samples and complex networks to process hard ones, thus reducing the average computational cost.
Generally, dynamic networks build a model ensemble through a cascaded~\cite{viola2004robust} or parallel~\cite{jacobs1991adaptive} structure and selectively activate the models conditioned on the input, allocating network with appropriate computational cost to each sample.
However, the vast network capacity and low inference cost of dynamic networks come at the cost of increased parameters~\cite{kong2021classsr,park2015big}, which makes it challenging to deploy to edge devices.

We note that pruning is a common technique that discards redundant components to obtain a slimmer network with comparable performance~\cite{frankle2018lottery,gao2018dynamic,oh2022attentive}, which can be 
categorized into unstructured pruning and structured pruning. 
Unstructured pruning removes individual weights at any location and can achieve a high compression ratio~\cite{han2015learning,guo2016dynamic}. 
However, the irregular sparsity patterns of weight tensors lead to poor utilization of current vector processing architectures~\cite{wen2016learning}.
By contrast, structured pruning removes predetermined structures, such as block~\cite{wang2019non} and kernel~\cite{tan2020pcnn,wen2016learning}, which can better utilize processor resources but fail to maintain accuracy.
Recently, $N$:$M$ fine-grained structured pruning~\cite{mishra2021accelerating,zhou2021learning} has emerged as a better alternative to combine both the advantage of unstructured pruning and structured pruning.
Here, the $N$:$M$ sparsity means that at most $N$ weights are non-zero for every continuous $M$ weight.
Such constraint allows the pruned network to be accelerated on modern hardware\cite{mishra2021accelerating}.
Moreover, we observe that high pruning ratio networks can still well restore lightly degraded images while highly degraded images require low pruning ratio networks to process, as shown in Figure~\ref{Fig.Observation}.
Therefore, we believe $N$:$M$ fine-grained structured pruning is a sweet solution to marry dynamic networks.

In this paper, we perform dynamic $N$:$M$ structured pruning on networks based on the input, aiming to reduce the average computational cost while introducing negligible parameters and maintaining performance.
We propose a new image restoration pipeline named ClassPruning, which assigns networks with different pruning ratios to images with varying restoration difficulties.
The pipeline consists of two parts: Classifier Module and Restoration Module.
The Classifier Module is a lightweight classification network that classifies the degraded image into a specific class based on its restoration difficulty. 
And the Restoration Module is an image restoration network on which we perform dynamic $N$:$M$ structured pruning to obtain the sparse networks with different capacities.
However, training a dynamic pruning network is not a trivial task, so we propose an improved training strategy.
We first introduce entropy loss and cost loss to enable the network to trade off performance and efficiency.
And we also use variable BatchNorm~\cite{ioffe2015batch} and SR-STE~\cite{zhou2021learning} to stabilize training and improve performance.
Finally, we propose a three-stage training scheme: we first pre-train the Classifier Module using pseudo labels, then fix the Classifier Module and only optimize the Restoration Module, and finally finetune the two parts simultaneously end-to-end until convergence. 
We conducted experiments on several representative image restoration networks, \emph{i.e.,} DnCNN~\cite{zhang2017beyond} for denoising, RESCAN~\cite{li2018recurrent} for deraining, SRN~\cite{tao2018scale} for deblurring and GCANet~\cite{chen2019gated} for dehazing. As shown in Figure~\ref{Fig.Front}, ClassPruning could help these restoration networks save approximately 40\% of computational cost while maintaining comparable performance.
\section{Related Work}

\noindent
\textbf{Image Restoration.}
In recent years, learning-based methods~\cite{zhang2017beyond,zamir2020learning,zamir2021multi,zhang2020residual,zamir2020cycleisp} are rapidly developing and achieving more promising results in various image restoration tasks.
DnCNN~\cite{zhang2017beyond} pioneered using convolution neural networks to handle different levels of Gaussian noise. 
RESCAN~\cite{li2018recurrent} builds a multi-stage architecture to obtain excellent deraining results. 
SRN~\cite{tao2018scale} explores a multi-scale network for boosting image deblurring performance.
And GCANet~\cite{chen2019gated} proposed a gated contextual aggregation network to predict the haze-free image. 
Despite the remarkable success, most models are complex and require huge computational resources, thus limiting their applications.  
So we do not aim to design a complicated network but propose a novel image restoration pipeline that reduces models' computational cost while introducing negligible parameters and maintaining performance via dynamic $N$:$M$ structured pruning.

\noindent
\textbf{Dynamic Networks.}
Compared to static models, which fix computational graphs and parameters once trained, dynamic networks adjust their structures or parameters to different inputs~\cite{wang2021not,yang2020resolution,yang2020mutualnet,li2021dynamic,chen2022arm}. 
Due to its property of assigning different capacity networks to varying inputs, researchers have built dynamic networks through dynamic depth~\cite{graves2016adaptive,wang2018skipnet}, dynamic width~\cite{li2021dynamic,yuan2020s2dnas} and dynamic routing~\cite{kong2021classsr,liu2018dynamic} to reduce the models' average computational cost.
Although these methods accelerate models, they often introduce many parameters, making models burdensome and unfriendly to devices with limited memory storage.
Therefore, in this paper, we are looking for a dynamic network that adjusts its network capacity based on the image restoration difficulties, which introduces negligible parameters and is hardware friendly.

\noindent
\textbf{Network Pruning.}
Removing unnecessary components of deep neural networks is a promising direction to compress and accelerate models~\cite{he2017channel,liu2019metapruning,you2019gate,he2021cap}.
In addition to structured pruning~\cite{han2015learning,frankle2018lottery,ramanujan2020s} and unstructured pruning~\cite{he2017channel,li2016pruning,liu2017learning}, recent works~\cite{mishra2021accelerating,zhou2021learning,oh2022attentive} have proposed $N$:$M$ fine-grained structured pruning as a better alternative that has high efficiency and lossless performance as well as is hardware friendly.
Nvidia~\cite{mishra2021accelerating} first presents the design and behavior of Sparse Tensor Cores supported by Ampere GPU architecture to exploit a $2$:$4$ sparsity pattern and describe a simple workflow for training sparse networks to maintain accuracy.
Then, SR-STE~\cite{zhou2021learning} is proposed to be a simple yet universal recipe for training a $N$:$M$ sparse neural network from scratch.
And SLS~\cite{oh2022attentive} introduce a layer-wise $N$:$M$ sparsity search framework to obtain an efficient image restoration network.
Unlike previous works that proposed a training strategy to obtain a high-performance static sparse network, this paper attempts to perform dynamic $N$:$M$ structured pruning on base restoration models to obtain sparse subnetworks with different capabilities for processing images with varying restoration difficulties.

\section{Method}
\subsection{Observation}

\begin{figure}[t]
    \centering
    \includegraphics[width=1.0\linewidth]{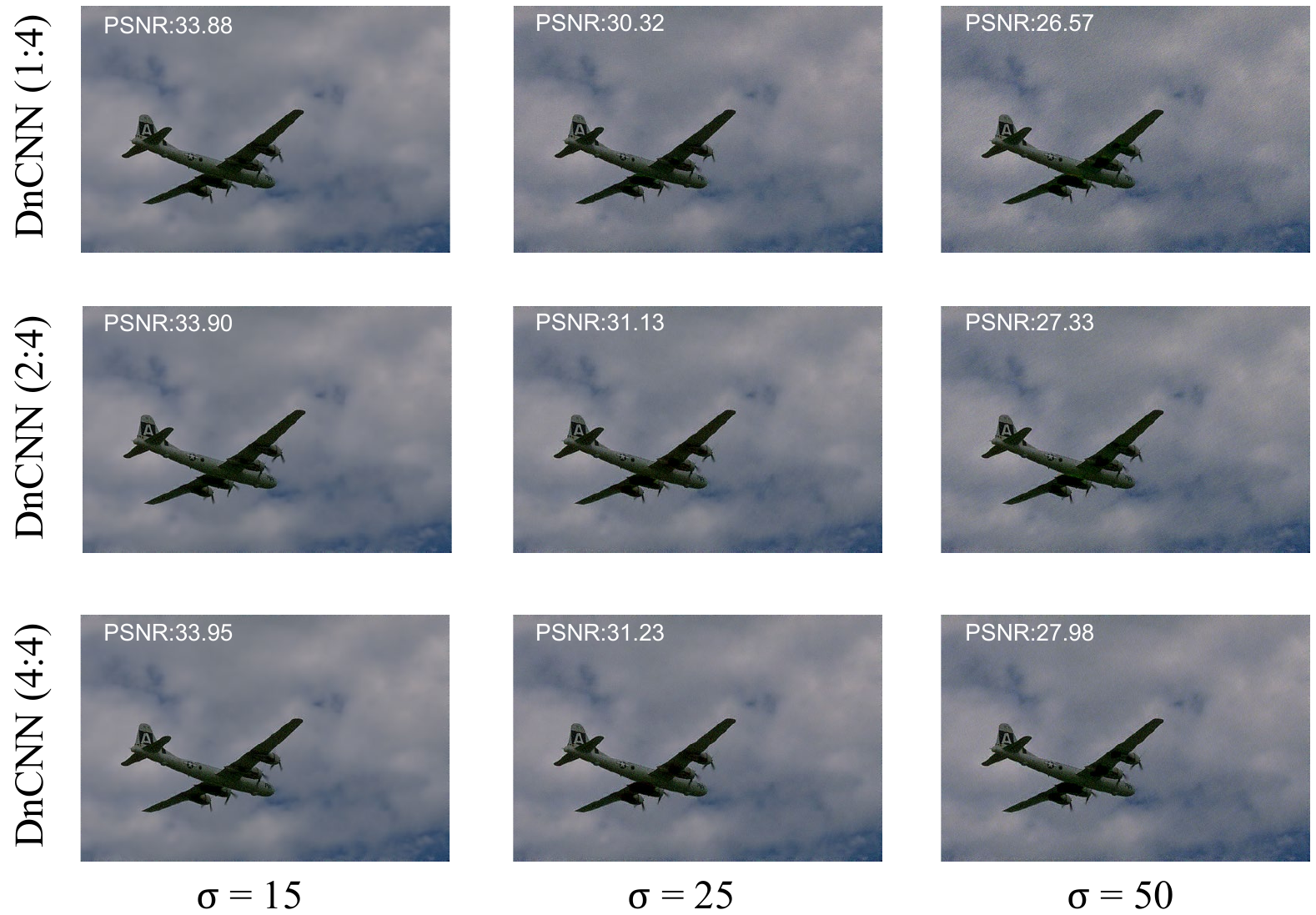}
    \centering
    \caption{PSNR and qualitative results of various noise level images obtained by sparse DnCNNs under different $N$:$M$ pruning ratios. Here we set the $N$:$M$ to be $1$:$4$, $2$:$4$ and $4$:$4$, as well as the noise levels $\sigma$ to 15, 25 and 50.
        }
    \vspace{0.5cm}
    \label{Fig.Observation}
\end{figure}

We first illustrate our observations on images with different restoration difficulties.
Specifically, we perform $N$:$M$ pruning~\cite{mishra2021accelerating} on DnCNN~\cite{zhang2017beyond} with different sparsity ratios to obtain varying capacity models for blind Gaussian denoising tasks.
From the results shown in Figure~\ref{Fig.Observation}, we find that images with low noise levels can still be restored well by DnCNN with a 75\% sparsity ratio. 
In contrast, only the unpruned model yields promising restored results for images with high noise levels.
Such a finding inspires us to preserve the whole ability of the network and perform $N$:$M$ pruning on the neural network dynamically according to restoration difficulties of input images, thus achieving a satisfying efficiency and performance tradeoff.
Note that such a method introduces only a few parameters so as not to lead to a burdensome model; instead, the $N$:$M$ fine-grained structured pruning makes the model hardware friendly.

\subsection{Network Architecture}
\begin{figure*}[t]
    \centering
    \includegraphics[width=0.9\textwidth]{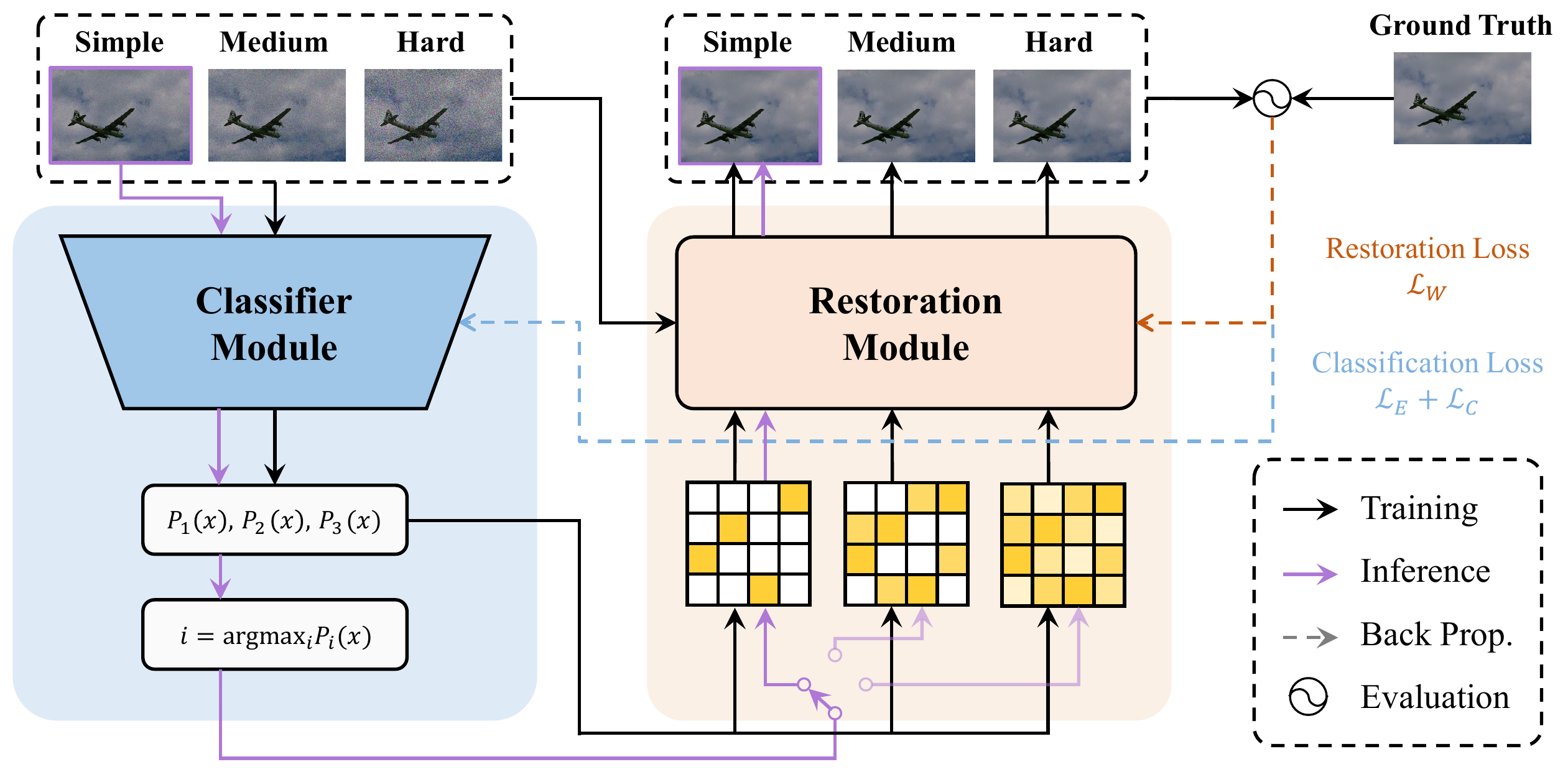}
    \centering
    \caption{The overview of the proposed ClassPruning, when the number of image restoration difficulty class $M$ = 3. Classifier Module: aims to generate the probability vector, Restoration Module: aims to restore the degraded images with corresponding sparse subnetworks.}
    \label{Fig.Arch}
\end{figure*}

Our proposed ClassPruning is a novel image restoration pipeline, as shown in Figure~\ref{Fig.Arch}, which consists of two parts: Classifier Module and Restoration Module.
The Classifier Module classifies the input into $L$ classes, while the Restoration Module can be an arbitrary image restoration network on which we perform $N$:$M$ fine-grained structured pruning with different sparsity ratios to obtain $L$ branches $\{f_{R}^{i}\}_{i=1}^L$ for dealing with different inputs.
Specifically, we first feed the degraded image $y$ into the Classifier Module to generate the probability vector $[p_1(y),..., p_L(y)]$ of restoration difficulty.
Then we determine which sparse subnetwork to be sampled based on the index of the maximum probability value $i = \arg{\max}_i  p_i(y)$.
Finally, we handle the input using the $i$-th sparse network of the Restoration Module.

\noindent
\textbf{Classifier Module.}
The Classifier Module aims to identify the restoration difficulty of the input degraded images efficiently.
Thus, we design a lightweight classification network to minimize the additional introduced computational cost and parameters, which contains four convolutional blocks (Conv-BatchNorm~\cite{ioffe2015batch}-ReLU~\cite{glorot2011deep}), an global average pooling layer~\cite{lin2013network}, and a fully-connected layer.
The convolutional blocks are responsible for image local features extraction, the global average pooling layer aggregates global information, and the fully-connected layer outputs the probability vector. 
And our experiments show that such a simple structure can yield satisfying results.

\noindent
\textbf{Restoration Module.}
The Restoration Module is designed to restore degraded images by sparse subnetworks with different capabilities based on the image restoration difficulty.
Unlike previous works~\cite{gao2018dynamic,lin2017runtime} of judging the importance of each convolutional kernel and conducting channel-wise pruning, we would like to prune networks at a fine-grained level to keep models' performance. 
Based on the prediction of the Classifier Module, we dynamically perform $N$:$M$ structured pruning with different sparsity ratios to obtain varying complexity subnetworks.   
We set the candidate pruning ratios to be $2$:$4$ and $1$:$4$, indicating 50\% and 75\% reduction in computational costs, respectively.
The two pruned networks, together with the base restoration network, form the three branches of the Restoration Module.
In this way, we can restore lightly degraded images with high pruning ratio networks and highly degraded images with low pruning ratio networks, thus speeding the image restoration model up while maintaining performance.

\subsection{Training Strategy}

\noindent
\textbf{Classifier Module.}
Most datasets for image restoration do not label the degraded images with restoration difficulty, preventing us from directly training the Classifier Module.
Therefore, we attempt to label the degraded images with the restoration difficulty class.
Considering that the degradation processes of tasks are different, we employ different methods to synthesize degraded images with restoration difficulty classes.
For image deblurring, we average the adjacent images $y_j$ and then determine the restoration difficulty class based on the number of fused images:
\begin{align}
    y_i = \sum_{j=1}^{N_i} y_{j},
\end{align}
where $N_i$ is the number of fused images for the $i$-th class of restoration difficulty.
And for denoising, deraining, and dehazing tasks, we first subtract the clean image $x$ from the degraded image $y$ to obtain the degraded components $R = y - x$.
Then we categorize all the $R$ into three classes according to their variance and map each $R$ to the median values of the three classes to satisfy the fact that the same scenario often has different levels of degradation:
\begin{align}
    R_i = \frac{R \sqrt{Var(R_{i}^{m})}}{\sqrt{Var[R]}},
\end{align}
where $i$ denotes the class and $m$ means the median value.
Finally, we add the new degraded component $R_i$ to the clean image to obtain the labeled degraded image $y_i$:
\begin{align}
    y_i = x + R_i.
\end{align}
So far, we have obtained degraded images with different restoration difficulties, and we can use Cross Entropy (CE) loss to optimize the Classifier Module.
However, apart from the degradation level, the restoration difficulty is also related to the content of the degraded images, so the labels obtained with the above method may be biased.
Therefore, we use an implicit training approach to finetune the Classifier Module.
Specifically, we propose entropy loss $L_E$ and cost loss $L_C$ to train the Classifier Module combined with the Restoration Module.
The entropy loss helps the Classifier to have much higher confidence in class with the maximum probability than others, which can accelerate the model convergence and ensure the stability of training:
\begin{align}
    L_E &= \sum_{i=1}^{L} - p_i(y) log(p_i(y)).
\end{align}
Besides, we aim to trade off the effectiveness and efficiency of restoration models. However, when only $L_E$ and reconstruction loss exist, all inputs will be classified as hard samples even though the Classifier Module attains a good initialization by CE loss.
To avoid such a bad case, we assign an overhead value $C_i$ to each branch and multiply it with the corresponding probability output by the Classifier Module to form the branch overhead. 
We sum all the overhead and minimize it to force the input to be classified into the most appropriate restoration difficulty class:
\begin{align}
    L_C &= \sum_{i=1}^{L} p_i(y) C_i.
\end{align}
Through subsequent training, we enable the Classifier Module to adaptively classify the images with different restoration difficulties and achieve satisfactory results.

\noindent
\textbf{Restoration Module.}
Our restoration network contains a base network and its sparse versions obtained by dynamic $N$:$M$ pruning. 
During testing, the input image only passes through one branch, which is determined by the index of the maximum probability value.
In contrast, during training, we allow all degraded images with different restoration difficulties to pass through all $L$ networks and adopt the weighted-$L_1$ loss $L_W$ to guide the models to reconstruct clean images, where the loss weights of each network are the probability output by the Classifier Module:
\begin{align}
    L_W = \sum_{i=1}^{L} p_i(y) ||f_R^i(y) - x ||_1.
\end{align}
And we weight $L_W$, $L_E$ and $L_C$ by $\omega_1$, $\omega_2$ and $\omega_3$ to form the final loss $L_F$:
\begin{align}
    L_F = \omega_1 L_W + \omega_2 L_E + \omega_3 L_C.
\end{align}
Also, we use some strategies to stabilize training and improve performance.

\textit{1) Variable BatchNorm.} The Restoration Module handles images of different restoration difficulties using sparse networks of various capacities.
However, when the network consists of BatchNorm layers, using the same BatchNorm layers for all sparse networks leads to severe performance drops due to different pruning ratios altering feature mean and variance. 
And we solve this problem by using the variable BatchNorm~\cite{li2021dynamic}, which assigns individual BatchNorm layers for different sparse networks, which are selectively activated depending on the predicted class.

\textit{2) SR-STE.} 
The pruning operation is non-differentiable, and researchers usually employ Straight-through Estimator (STE)~\cite{bengio2013estimating} to compensate for sparse neural networks' backpropagation.
However, STE only updates the weights that are not set to zero, which might harm our networks' performance since all our subnetworks are ensembled in one model, and individual weight updates are inappropriate.
So we use SR-STE~\cite{zhou2021learning}, which updates all parameters by adding a regularization term to decrease the magnitude of the pruned weights.
Experiments show that SR-STE can further improve the model performance.

\noindent
\textbf{Training Pipeline.}
The most intuitive training method is to train the Classifier Module and Restoration Module together from scratch. 
However, we find this strategy is quite unstable, and even if this model converges, it does not produce promising results.
Thus, we propose to train the model in three stages:
\begin{itemize}
    \item \textbf{Stage 1}: Pretrain the Classifier Module with CE loss using the degraded images with pseudo labels;
    \item \textbf{Stage 2}: Fix the parameters of the Classifier Module and train the Restoration Module using $L_W$;
    \item \textbf{Stage 3}: Relax all parameters and finetune the whole model with the final loss $L_F$.
\end{itemize}
Experimental results show that this scheme can stabilize training and prevent the network from falling into trivial solutions, and somehow boost the model's performance.

\subsection{Implementation Details}
Considering that we obtain the model in three stages, we first train the Classifier Module using the degraded images with pseudo labels. 
The mini-batch size is set to 32, and CE Loss is adopted with Adam optimizer~\cite{kingma2014adam}.
And we use the cosine annealing learning strategy~\cite{loshchilov2016sgdr} to adjust the learning rate. 
The total epoch is $200$ with the learning rate from $10^{-3}$ to $10^{-6}$.
Then, we fix the Classifier Module and optimize the Restoration Module using $L_W$ with Adam optimizer. 
The training epoch is $500$ with the learning rate from $10^{-4}$ to $10^{-6}$.
In the last stage, we relax all the parameters and train the whole model using $L_F$ with Adam optimizer, and the learning rate decays from $10^{-5}$ to $10^{-7}$ for $500$ epochs.
For fair comparisons, we train all the base models for $1,000$ epochs with the learning rate from $10^{-4}$ to $10^{-7}$.
More details on training protocols are presented in the supplementary material.

\section{Experiment}
We evaluate our proposed ClassPruning on four different image restoration tasks: (\textbf{a}) image denoising, (\textbf{b}) image deraining, (\textbf{c}) image deblurring, and (\textbf{d}) image dehazing.
\subsection{Datasets}

\noindent
\textbf{Image Denoising.} For gaussian image denoising, we follow ~\cite{chen2016trainable,zhang2017beyond} to use 400 images of size 180 $\times$ 180 for training and BSD68~\cite{roth2009fields} for testing. 
Noisy images are generated by adding additive white Gaussian noise with noise level $\sigma$ to clean images, and we set $\sigma$ to be 15, 25, and 50 for both training and testing.
And for real image denoising, we use 320 high-resolution images of the SIDD dataset~\cite{abdelhamed2018high} to train our model and perform the evaluation on 1280 patches from the SIDD validation set. 

\noindent
\textbf{Image Deraining.} Following ~\cite{li2018recurrent}, we select 700 images from Rain800~\cite{zhang2019image} to train our model and perform the evaluation on the remaining 100 images, named Test100, for testing. Also, we use Rain100H~\cite{yang2017deep} as the testing dataset to evaluate our model. 
Note that the PSNR/SSIM scores are computed on the RGB channels instead of the Y channel.

\noindent
\textbf{Image Deblurring.} Consistent with prior works~\cite{nah2017deep,tao2018scale}, we use the GoPro~\cite{nah2017deep} dataset for training our model. It contains 2,103 blurry sharp image pairs for training and 1,111 for validation. We also use Kohler dataset~\cite{kohler2012recording} for evaluation, which consists of 4 images, and each image is blurred with 12 different kernels.

\noindent
\textbf{Image Dehazing.} Prior work~\cite{li2018benchmarking} proposes an image dehazing benchmark RESIDE, which consists of large-scale training and testing hazy image pairs synthesized from depth and stereo datasets. 
We leverage the RESIDE's Indoor Training Set for training, which contains 1399 clean images and 13990 hazy images generated by corresponding clean images. 
And we evaluate the models' performance on Synthetic Objective Testing Set (SOTS), which contains 500 indoor images and 500 outdoor ones.

\subsection{Comparison with Base Models}

\begin{table*}[t]
  \caption{Comparison between the base models and ClassPruning networks on denoising, deraining, deblurring, and dehazing tasks. The CP-$Model$ means the model is obtained using our proposed ClassPruning, where $Model$ corresponds to the specific model name.}
    \centering    
    \resizebox{\linewidth}{!}{
    \small
    \begin{tabular}{lccccccccc}
        \hline
        \multirow{2}*{{Methods}}&\multicolumn{4}{c}{{\hspace{-0.75cm} BSD68}} & &\multicolumn{4}{c}{\hspace{-0.75cm} {SIDD}}\\
        \cline{2-5}
        \cline{7-10}
        & {$\sigma=15$} ~~~~ {$\sigma=25$}&{$\sigma=50$} & {FLOPs} & \#Params & &{PSNR}&~~~~~~~{SSIM}& {FLOPs} & \#Params\\
        \hline
        DnCNN & 31.63 ~~~~~~~ 29.18 & 26.24 &87.28G (100\%) &0.668M & &38.43 &~~~~~~~0.910 &87.58G (100\%)& 0.671M\\
        CP-DnCNN& 31.59 ~~~~~~~ 29.17 &26.26 &50.88G (58.3\%) &0.754M & &38.38 &~~~~~~~0.908 &53.33G (60.9\%) &0.758M \\
        \bottomrule
        \toprule
        \multirow{2}*{{Methods}}&\multicolumn{4}{c}{\hspace{-0.75cm}{Test100}} & &\multicolumn{4}{c}{\hspace{-0.75cm}{Rain100H}}\\
        \cline{2-5}
        \cline{7-10}
        & \hspace{-1.3cm} {PSNR}&\hspace{-1cm} {SSIM} & {FLOPs} & \#Params & &{PSNR}&~~~~~~~{SSIM} & {FLOPs} & \#Params\\
        \hline
        RESCAN & \hspace{-1.3cm}24.46 & \hspace{-1cm}0.838 & 64.50G (100\%)&0.589M & &25.71 &~~~~~~~0.832 &64.50G (100\%)&0.589M\\
        CP-RESCAN& \hspace{-1.3cm}24.38 &\hspace{-1cm}0.838 &36.18G (56.1\%)&0.672M & &25.60 &~~~~~~~0.830 &34.31G (53.2\%)&0.672M\\
        \bottomrule
        \toprule
        \multirow{2}*{{Methods}}&\multicolumn{4}{c}{\hspace{-0.75cm}{GoPro}} & &\multicolumn{4}{c}{\hspace{-0.75cm} {Kohler}}\\
        \cline{2-5}
        \cline{7-10}
        & \hspace{-1.3cm} {PSNR}& \hspace{-1cm}{SSIM} & {FLOPs}& \#Params & &{PSNR}&~~~~~~~{SSIM} & {FLOPs}& \#Params\\
        \hline
        SRN & \hspace{-1.3cm} 30.21 & \hspace{-1cm}0.933& 217.2 (100\%)&10.25M & &26.70 &~~~~~~~0.835 &217.2 (100\%)&10.25M\\
        CP-SRN& \hspace{-1.3cm} 30.18 & \hspace{-1cm}0.932& 131.8 (60.7\%)&10.33M & & 26.71 &~~~~~~~0.834 &132.7 (61.1\%)&10.33M\\
        \bottomrule
        \toprule
        \multirow{2}*{{Methods}}&\multicolumn{4}{c}{\hspace{-0.75cm}{RESIDE-IN}} & &\multicolumn{4}{c}{\hspace{-0.75cm}{RESIDE-OUT}}\\
        \cline{2-5}
        \cline{7-10}
        & \hspace{-1.3cm} {PSNR}& \hspace{-1cm}{SSIM} & {FLOPs}& \#Params & &{PSNR}&~~~~~~~{SSIM} & {FLOPs}& \#Params\\
        \hline
        GCANet & \hspace{-1.3cm} 30.21 & \hspace{-1cm} 0.979& 36.82 (100\%)&0.702M & & 30.36  &~~~~~~~0.963 &36.82 (100\%)&0.702M\\
        CP-GCANet& \hspace{-1.3cm} 30.17 & \hspace{-1cm}0.977& 22.24 (60.4\%)&0.785M & & 30.35 &~~~~~~~0.959 &21.68 (58.9\%)&0.785M\\
        \bottomrule
    \end{tabular}
    }
    \label{tab:Compare}
\end{table*}

To demonstrate the effectiveness and generalization of the method, we use ClassPruning to speed up some representative networks for various image restoration tasks.
Specifically, we use DnCNN~\cite{zhang2017beyond}, RESCAN~\cite{li2018recurrent}, SRN~\cite{tao2018scale} and GCANet~\cite{chen2019gated} for denoising, deraining, deblurring and dehazing, respectively. 
Training and testing follow the strategy described above, and results are summarized in Table~\ref{tab:Compare}.
In summary, ClassPruning helps models obtain comparable performance to the base networks and accelerate all base models approximately 1.67 times (40\% FLOPs reduction) by introducing only a few parameters, mainly belonging to the Classifier Module.
Also, we find that the reduction of FLOPs is correlated with the restoration task and the test dataset.
With comparable performance, the Kohler dataset in the deblurring task is the most difficult, reducing only 39\% FLOPs, while the Rain100H dataset in the deraining task is the easiest, reducing 45\% FLOPs.

\subsection{Comparison with Other Dynamic Networks}

\begin{table}[t]
  \caption{Comparison with different dynamic approaches. Multi-Width and Multi-Depth are formed by multiple independent networks with different widths and depths. Channel-Skip and Layer-Skip are obtained by skipping layers and channels of the network.}
  \centering
  \resizebox{\linewidth}{!}{
  \begin{tabular}{lccccc}
    \hline
      \multirow{2}*{{Method}}&\multicolumn{5}{c}{BSD68}\\
      \cline{2-6}
      &{$\sigma=15$} & {$\sigma=25$} & {$\sigma=50$}& {FLOPs} & {\#Params}\\
    \hline
    Multi-Width~\cite{eigen2013learning} & 31.47& 29.15& 26.23 & 50.77G& 1.251M\\
    Multi-Depth~\cite{park2015big} & 31.60 & 29.17 & 26.27 & 52.11G&1.272M\\
    Channel-Skip~\cite{li2021dynamic}  &31.38 & 29.03 & 26.21 &50.71G&0.754M\\
    Layer-Skip~\cite{wu2018blockdrop} &31.42 & 29.08 & 26.24 &51.58G&0.754M\\
    Ours &31.59 &29.17 &26.26 &50.88G& 0.754M\\
    \bottomrule
  \end{tabular}
  }
  \label{tab:Dynamic Ablation}
\end{table}

We compare ClassPruning with other dynamic networks on the denoising task to demonstrate the effectiveness of our method.
And here, we adopted four different dynamic networks: Multi-Width~\cite{eigen2013learning}, Multi-Depth~\cite{park2015big}, Channel-Skip~\cite{li2021dynamic} and Layer-Skip~\cite{wu2018blockdrop}.
The first two selectively execute one of the multiple independent candidate networks with different depths and widths, respectively. 
And the last two skip the execution of intermediate layers and channels, respectively.
The results are shown in Tabel~\ref{tab:Dynamic Ablation}, and we can see that our method outperforms most dynamic approaches, especially the channel-skip, which performs channel-wise structured pruning.
We also found that dynamic networks ensembling all the subnetworks in one model are more challenging to optimize than independent subnetworks, and our training strategy could be a good solution.
Although our method is slightly inferior to the multi-depth dynamic approach, it is worth noting that ClassPruning introduces fewer parameters, \emph{i.e.} more lightweight, and the $N$:$M$ sparsity pattern makes our method more friendly to hardware.

\subsection{Analysis}

\begin{figure*}[t]
  \centering
  \includegraphics[width=1.0\textwidth]{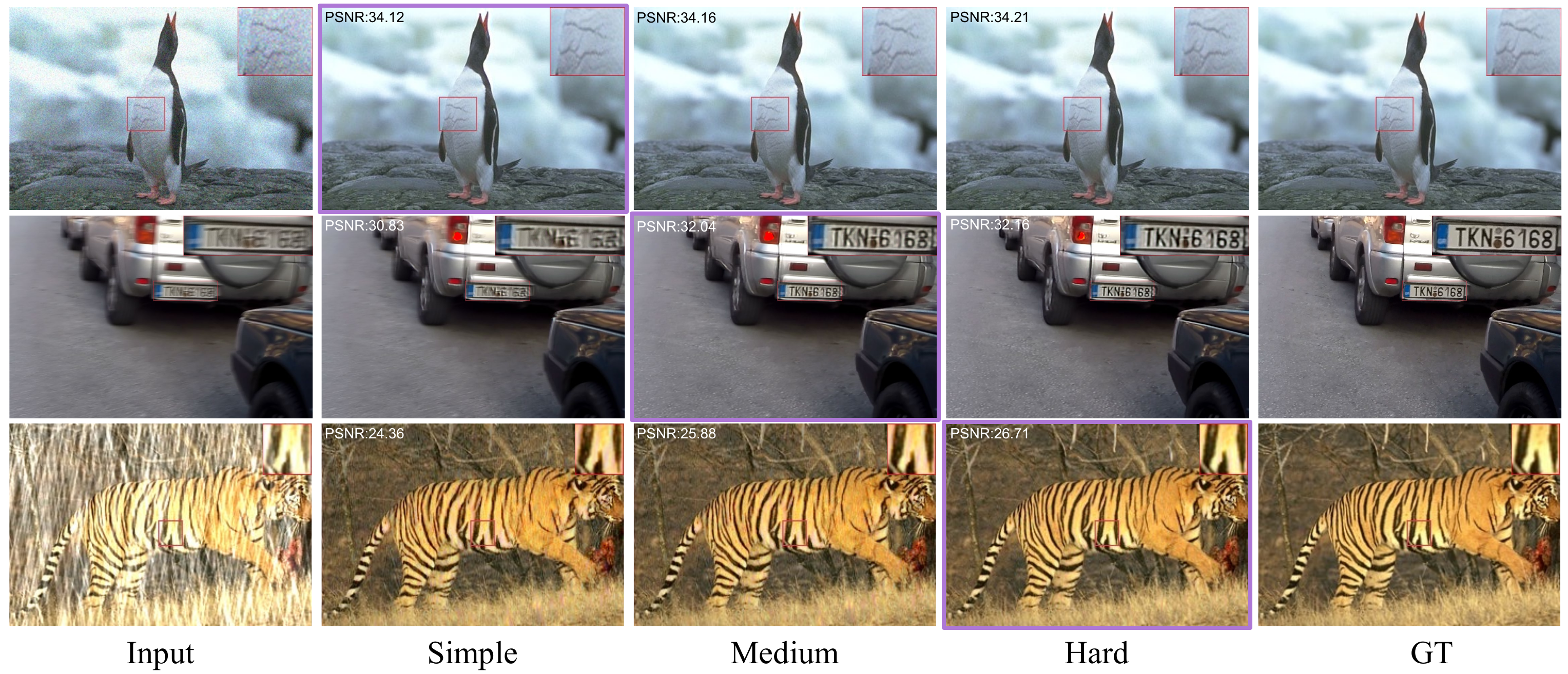}
  \centering
  \caption{Visual samples obtained by sparse networks with different capabilities. The purple outer box corresponds to the difficulty class predicted by our Classifier Module.}
  \label{Fig.Result}
\end{figure*}

\noindent
\textbf{Visualization.}
Figure~\ref{Fig.Result} shows some samples of results obtained by sparse models with different capabilities, and the purple box indicates the branch selected by our Classifier Module.
We can see that lightly degraded images can be restored well by sparse networks with high pruning ratios, and only minor improvements are achieved with more complex models.
However, when we employ sparse networks with high pruning ratios to handle highly degraded images, the performance drops dramatically and unsatisfying results are produced.
Also, it can be seen that our method appropriately assigns networks with different capacities to images with varying restoration difficulties, thus reducing the average computational cost while maintaining performance.

\begin{table}[t]
  \caption{Comparison between training from scratch ($^\dagger $) and training with our proposed strategy. Note that we train from scratch multiple times and select the best results due to its instability.}
  \centering
  \resizebox{\linewidth}{!}{
  \large
  \begin{tabular}{lccc}
      \hline
      \multirow{2}*{{Methods}}&\multicolumn{3}{c}{\hspace{-1.3cm}{BSD68}}\\
      \cline{2-4}
      & {$\sigma=15$} ~~ {$\sigma=25$}&{$\sigma=50$} & {FLOPs}\\
      \hline
      CP-DnCNN$^\dagger $&31.43 ~~~~ 29.01&26.08 &51.06G (58.5\%)  \\
      CP-DnCNN& 31.59 ~~~~ 29.17 &26.26 &50.88G (58.3\%) \\
      \bottomrule
      \toprule
      \multirow{2}*{{Methods}}&\multicolumn{3}{c}{\hspace{-1.3cm}{Test100}}\\
      \cline{2-4}
      & \hspace{-1.3cm}{PSNR}& \hspace{-1.3cm} {SSIM} & {FLOPs}\\
      \hline
      CP-RESCAN$^\dagger $& \hspace{-1.3cm} 24.30 &\hspace{-1.3cm} 0.835&37.47G (58.1\%)\\
      CP-RESCAN& \hspace{-1.3cm} 24.38 &\hspace{-1.3cm} 0.838 &36.18G (56.1\%)\\
      \bottomrule
      \toprule
      \multirow{2}*{{Methods}}&\multicolumn{3}{c}{\hspace{-1.3cm} {GoPro}}\\
      \cline{2-4}
      & \hspace{-1.3cm} {PSNR}&\hspace{-1.3cm}  {SSIM} & {FLOPs}\\
      \hline
      CP-SRN$^\dagger $& \hspace{-1.3cm} 30.05 & \hspace{-1.3cm} 0.928& 130.01 (59.9\%) \\
      CP-SRN& \hspace{-1.3cm} 30.18 &\hspace{-1.3cm}  0.932& 131.83 (60.7\%)\\
      \bottomrule
      \toprule
      \multirow{2}*{{Methods}}&\multicolumn{3}{c}{\hspace{-1.3cm} {Indoor}}\\
      \cline{2-4}
      & \hspace{-1.3cm} {PSNR}&\hspace{-1.3cm}  {SSIM} & {FLOPs}\\
      \hline
      CP-GCANet$^\dagger $&\hspace{-1.3cm} 30.02 & \hspace{-1.3cm} 0.970 &21.94 (59.6\%)\\
      CP-GCANet&\hspace{-1.3cm} 30.17 &\hspace{-1.3cm}  0.977& 22.24 (60.4\%)\\
      \bottomrule
  \end{tabular}
  }
  \label{tab:Strategy}
\end{table}

\noindent
\textbf{Training Strategy.}
We obtained the restoration models by training with our proposed strategy and from scratch, respectively, to explore the impact of the training strategy on various tasks.
It is worth noting that training from scratch is quite unstable and sensitive to the weight ratio of $L_C$ and $L_W$, leading the Classifier Module to classify all inputs into one restoration difficulty easily; here, we train from scratch multiple times and select the best results.
Table~\ref{tab:Strategy} shows that our strategy can trade off performance and efficiency better, especially in denoising and deraining tasks.
Also, the novel strategy enables our training to be more stable and alleviates the dilemma of being sensitive to losses.

\begin{table}[t]
  \caption{Ablation experiments with two optimization methods, variable BatchNorm and SR-STE, on the BSD68 dataset.}
  \centering
  \resizebox{\linewidth}{!}{
  \begin{tabular}{ccccc}
    \hline
    \multirow{2}*{{Method}}&\multicolumn{4}{c}{{BSD68}}\\
    \cline{2-5}
    &{$\sigma=15$} &{$\sigma=25$} &{$\sigma=50$} & {FLOPs}\\
    \hline
    vBN $\rightarrow$ BN& 30.02& 28.14& 25.58& 59.69G (68.4\%)\\
    SR-STE $\rightarrow$ STE&31.40& 28.91& 25.94&62.23G (71.3\%)\\
    Ours  &31.59&29.17&26.26&50.88G (58.3\%)\\
    \hline
  \end{tabular}
  }
  \label{tab:tricks}
\end{table}

\noindent
\textbf{Variable BN \& SR-STE.}
As mentioned earlier, we employ variable BatchNorm and SR-STE, and here we conducted ablation experiments to investigate their effects as shown in Table~\ref{tab:tricks}.
The performance drops severely with the same BatchNorm layers for all sparse networks.
We believe this is because pruning leads to different statistics of features and variable BatchNorm provides a solution.
Also, STE is inferior to SR-STE, which only updates non-zero weights that would harm other sparse networks.

\begin{figure*}[t]
  \centering
  \includegraphics[width=1.0\textwidth]{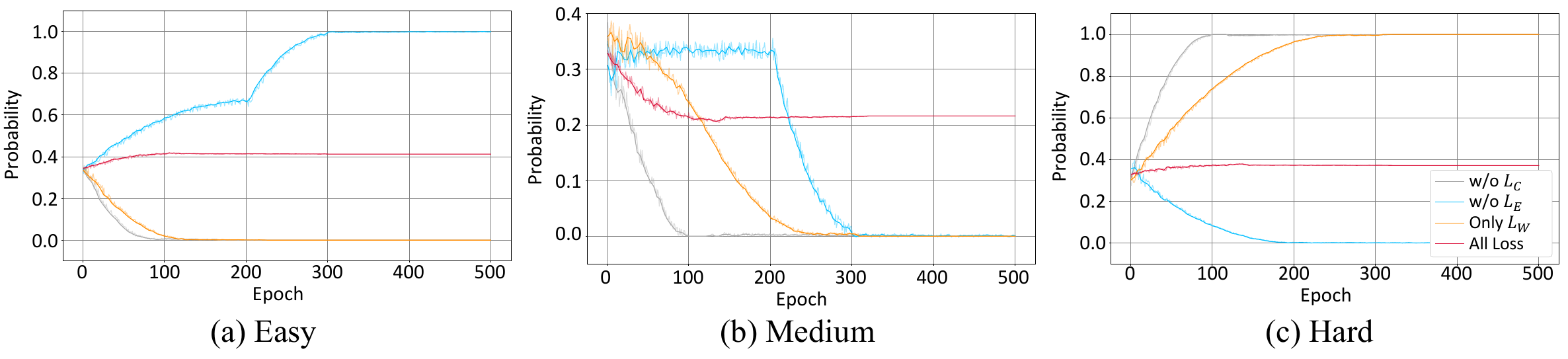}
  \centering
  \caption{Training probability curves comparison of ClassPruning with various losses on the BSD68 dataset.}
  \label{Fig.Prob}
\end{figure*}

\noindent
\textbf{Loss Function.}
We further investigate the effect of the cost loss $L_C$ and entropy loss $L_E$ by removing them from the loss function. 
For simplicity, we present the restoration difficulty classification output by the Classifier Module under different losses on BSD68 in Figure~\ref{Fig.Prob}.
We can see that, without $L_C$, the Classifier Module converges quickly; however, it assigns all degraded images to the hard class, which is a bad local minimum for optimization.
And when we remove the $L_E$, the Classifier Module still classifies all inputs into one class, but the classified class is sensitive to the weight ratios of $L_W$ and $L_C$.
Here we present the probability curve by directly removing the $L_E$ from the baseline.
We also only use $L_W$ to train our model and find that the probability curve is similar to the loss without $L_C$ but converges more slowly, indicating our proposed $L_E$ could help the model converge faster.
And only when the three losses are used together, the Classifier Module can classify input images appropriately in a more stable way.

\begin{table}[t]
  \caption{Ablation experiments of performance and computational cost with different network widths on the denoising task.}
  \centering
  \resizebox{\linewidth}{!}{
  \huge
  \begin{tabular}{lcccccc}
    \hline
    \multirow{2}*{{Width}}&\multirow{2}*{{Model}}&\multicolumn{2}{c}{{BSD68}}&&\multicolumn{2}{c}{{SIDD}}\\
    \cline{3-4}
    \cline{6-7}
    & &{PSNR} & {FLOPs} & &{PSNR} &{FLOPs}\\
    \hline
    \multirow{2}*{16}&{DnCNN} & 27.77 & 5.520G(100\%)  &  &37.08 &5.580G(100\%)\\
    &{CP-DnCNN}& 27.68 & 3.320G (60.2\%)& & 35.62&3.930G (69.1\%)\\    
    \hline
    \multirow{2}*{32}&{DnCNN} & 28.31 & 21.90G (100\%) & &37.54 &22.04G (100\%)\\
    &{CP-DnCNN}& 28.24& 13.01G (59.4\%)& &37.06  &14.37G (63.8\%)\\    
    \hline
    \multirow{2}*{64}&{DnCNN} & 29.01 & 87.28G (100\%)  & & 38.43  &87.58G (100\%)\\
    &{CP-DnCNN}& 29.00 & 50.88G (58.3\%)&  & 38.38  &53.33G (60.9\%)\\    
    \hline
    \multirow{2}*{128}&{DnCNN} & 29.32& 348.5G (100\%) & &38.77  &349.1G (100\%)\\
    &{CP-DnCNN}& 29.34& 202.5G (58.1\%)& &38.78  &208.8G (59.8\%)\\    
    \hline
    \multirow{2}*{256}&{DnCNN} & 29.52& 1,393G (100\%) & &38.90  &1,394G (100\%)\\
    &{CP-DnCNN}& 29.55& 806.4G (57.9\%)& &38.87 &812.7G (58.3\%)\\    
    \hline
  \end{tabular}
  }
  \label{tab:channelAblation}
\end{table}

\noindent
\textbf{Network Width.}
Considering that we perform $N$:$M$ pruning on the convolution kernel to obtain the different capacity sparse networks, it is essential to evaluate the impact of the number of the convolution kernel \emph{i.e.}, network channel.
We conducted the network width ablation study on the denoising task, and Table~\ref{tab:channelAblation} shows the performance and computational cost comparisons of models with different widths.
When the network width is large enough to be over-parameterized, our method can filter the network redundancy sufficiently, reducing more than 40\% computational cost while maintaining or exceeding the original DnCNN's performance.
Such a property is quite valuable since many models nowadays are incredibly complex, and our method can effectively speed them up.

\begin{table}[t]
  \caption{Ablation experiments with different restoration difficulty classifications on the BSD68 dataset. We use \textbf{1, 2, 4} to represent the $1$:$4$, $2$:$4$ and $4$:$4$ fine-grained structured pruning respectively.}
  \centering
  \resizebox{\linewidth}{!}{
  \normalsize
  \begin{tabular}{lccccc}
    \hline
      \multirow{2}*{{Type}}&\multicolumn{5}{c}{{BSD68}}\\
      \cline{2-6}
      &{$\sigma=15$} & {$\sigma=25$} & {$\sigma=50$}& Avg& {FLOPs}\\
    \hline
    \textbf{1} & 31.21& 28.69& 25.83& 28.58 &21.82G (25.0\%)\\
    \textbf{2} & 31.43& 28.85& 25.98& 28.75 &43.64G (50.0\%)\\
    \textbf{4} & 31.63& 29.18& 26.24& 29.02 &87.28G (100\%)\\
    \textbf{1\&2} & 31.28 & 28.74 & 25.88 &28.63&29.33G (33.6\%)\\
    \textbf{1\&4} & 31.39 & 28.89 & 26.00 & 28.76 &49.84G (57.1\%)\\
    \textbf{2\&4} & 31.62 & 29.18 & 26.26 &29.02&65.54G (75.1\%)\\
    \textbf{1\&2\&4} &31.59 &29.17 &26.26 &29.01&50.88G (58.3\%)\\
    \bottomrule
  \end{tabular}
  }
  \label{tab:numberAblation}
\end{table}

\noindent
\textbf{Classification Number.}
An essential concern in ClassPruning is choosing the appropriate classification number to optimally achieve a balance between performance and efficiency.
And we conducted experiments to compare several different classification types.
Here we name the \textbf{1, 2, 4} to represent the $1$:$4$, $2$:$4$ and $4$:$4$ fine-grained structured pruning, and the results are shown in table~\ref{tab:numberAblation}.
We find that when the pruning ratio is relatively high or only one kind of pruning exists, the restoration models' performance drops severely (\emph{e.g.}, \textbf{1}, \textbf{2} and \textbf{1\&2}).
However, when we use \textbf{2\&4} as our classification type, the model performs well, but the computational cost reduction is not attractive enough.
Therefore, we choose the classification type, which divides the restoration difficulty into three classes to trade off the computational cost and performance.

\section{Conclusion}
This paper proposes a general network slimming pipeline named ClassPruning, which could speed up existing learning-based image restoration models by approximately 1.67 times while maintaining performance.
The key idea is first to use the Classifier Module to determine the restoration difficulties of the input images.
And then, we perform dynamic $N$:$M$ fine-grained structured pruning on the restoration model based on the predicted restoration difficulty, thus obtaining the corresponding sparse subnetworks to process the input images.
Further, we propose a novel training strategy and two additional loss terms to stabilize training and improve performance.
Extensive experiments on various tasks and datasets demonstrate that our ClassPruning could save approximately 40\% FLOPs while introducing only a few parameters, indicating the effectiveness and generalization of ClassPruning.

{\small
\bibliographystyle{ieee_fullname}
\bibliography{egbib}
}

\end{document}